\documentclass[runningheads]{llncs}
\usepackage[T1]{fontenc}
\usepackage{graphicx}
\usepackage[ruled,lined,boxed,commentsnumbered]{algorithm2e}
\usepackage{microtype}
\usepackage{multirow}
\usepackage{multicol}
\usepackage{amsmath}
\usepackage{booktabs,subcaption,amsfonts,dcolumn}
\usepackage{color} 
\usepackage{url}
\usepackage{hyperref}
\usepackage{graphicx}
\usepackage{booktabs}
\usepackage{multirow}
\usepackage{caption}
\usepackage{float}
\usepackage{longtable}
\usepackage{bbding}
\usepackage{wrapfig}
\SetKwComment{Comment}{/* }{ */}
\usepackage{xcolor}
\usepackage[misc]{ifsym}

\definecolor{Mycolor1}{HTML}{BAD8F2}
\definecolor{Mycolor2}{HTML}{E8F2FB}
\definecolor{Mycolor3}{HTML}{FAE4E3}
\definecolor{reds}{RGB}{184,84,80}
\definecolor{blues}{RGB}{16,115,158}

\begin{document}
\title{Beyond Isolation: Multi-Agent Synergy for Improving Knowledge Graph Construction}
\titlerunning{Abbreviated paper title}
%
\author{Hongbin Ye \inst{1} $^{(\textrm{\Letter})}$\and
Honghao	Gui \inst{2} \and
Aijia Zhang \inst{1}\and
Tong Liu \inst{1} \and
Weiqiang Jia \inst{1}
}
%
\authorrunning{H. Ye et al.}
%

\institute{Zhejiang Lab, Hangzhou \and
Ant Group, Hangzhou 
}

\maketitle              
\begin{abstract}
This paper introduces CooperKGC, a novel framework challenging the conventional solitary approach of large language models (LLMs) in knowledge graph construction (KGC).  CooperKGC establishes a collaborative processing network, assembling a team capable of concurrently addressing entity, relation, and event extraction tasks.  Experimentation demonstrates that fostering collaboration within CooperKGC enhances knowledge selection, correction, and aggregation capabilities across multiple rounds of interactions.

\end{abstract}
\section{Introduction}

In the era of information abundance, constructing comprehensive knowledge graphs \cite{DBLP:conf/emnlp/Ye0CC22,DBLP:conf/acl/MondalHJ21,DBLP:conf/www/VakajTMOM23} has emerged as a pivotal task.
The advent of LLMs, such as GPT-3 \cite{DBLP:conf/nips/BrownMRSKDNSSAA20} and ChatGLM \cite{DBLP:conf/acl/DuQLDQY022}, has revolutionized natural language processing by showcasing unparalleled proficiency in understanding and generating human-like text.
However, the application of these models to KGC remains an intricate challenge, as this task necessitates not only language understanding but also precise extraction of elements within the confines of predefined schemas. 
Recent investigations \cite{DBLP:journals/corr/abs-2302-10205} reveals that the raw textual data utilized to train large language models may lack task-specific schemas, resulting in a weakened semantic grasp and structural analysis of the underlying schema.  Therefore we contends that a shift from traditional parameter-based paradigms to a more nuanced approach, like \emph{Chain-of-Thought} (CoT) \cite{DBLP:conf/nips/Wei0SBIXCLZ22,DBLP:journals/corr/abs-2310-04959}, can address the challenges posed by multi-step inference problems inherent in KGC. 
Embracing the profound insights from the \emph{Society of Mind} (SOM) \cite{minsky1988society}, which conceptualizes the mind as a complex system emerging from the interactions of simple components,
our research explores the transformative potential of LLM-based agents in multi-agent systems. 
Taking inspiration from pioneering work of \cite{DBLP:journals/corr/abs-2305-19118}, we employ the multi-agent debate framework for collaborative self-reflection on challenging tasks. 
Collaboration is defined as an iterative refinement process, wherein each round generates a new answer based on prior answers and self-reflection. 
This iterative feedback fosters continuous improvement, making our collaborative approach adept at tackling problems that elude single-agent solutions. 

Specifically, our dedicated team of agents comprises experts proficient in various tasks, including named entity recognition, relation extraction, and event extraction.
In our approach \textbf{CooperKGC} , we construct a collaborative team of agents, each specializing in distinct tasks to simulate the nuanced teamwork prevalent in human society. 
The integration of open interaction, expertise refinement, and adaptability to others' opinions mirrors the foundations of a cohesive society.
Our exploration into diverse collaboration strategies reveals key insights: 
(1) Inclusion of agents with varied expertise enhances collaboration outcomes.
(2) While model hallucinations \cite{DBLP:journals/corr/abs-2309-06794} may arise, effective communication among team members mitigates these drawbacks.
(3) Substantial team collaboration enhances extraction results on target tasks;  however, an intriguing observation emerges that increasing cooperation rounds doesn't invariably yield superior results.  In our collaborative mechanism, balancing interaction frequency ensures the expert agent's beliefs remain undisturbed by excessive external authoritative information, aligning with fundamental theories of sociology \cite{tuckman1965developmental,goffman2002presentation,durkheim2018division}.

\section{Related work}

\newcommand{\addcheckemoji}{\raisebox{-.2\height}{\includegraphics[height=12pt]{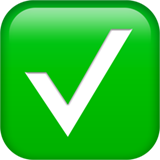}}}
\newcommand{\addcrossemoji}{\raisebox{-.2\height}{\includegraphics[height=12pt]{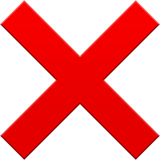}}}

\begin{table*}[t]

\resizebox{0.96\textwidth}{!}{
\begin{tabular}{@{}lccccc@{}}
    \toprule 
    & \textbf{\begin{tabular}[c]{@{}c@{}}Has multiple agents \\ involved? \end{tabular}} 
    & \textbf{\begin{tabular}[c]{@{}c@{}}Has personalized \\ agents? \end{tabular}} 
    & \textbf{\begin{tabular}[c]{@{}c@{}} Has interactive \\ rounds?\end{tabular}}
    & \textbf{\begin{tabular}[c]{@{}c@{}}Involves chain of thought \\ processes? \end{tabular}} 
    & \textbf{\begin{tabular}[c]{@{}c@{}}Accomplishes multiple \\ tasks in parallel? \end{tabular}} \\
    \midrule
    
    AutoKG~\cite{DBLP:journals/corr/abs-2305-13168}    &\addcrossemoji{}    & \addcrossemoji{} &  \addcrossemoji{}
    &\addcrossemoji{}    & \addcrossemoji{}  \vspace{1pt}\\

    ChatIE~\cite{DBLP:journals/corr/abs-2302-10205}    &\addcheckemoji{}  (2-agents)   & \addcrossemoji{} &  \addcheckemoji{} (2-rounds)
    &\addcrossemoji   & \addcrossemoji{}  \vspace{1pt}\\

    GPT-NER~\cite{DBLP:journals/corr/abs-2304-10428}    &\addcrossemoji{}    & \addcheckemoji{} &  \addcheckemoji{} (>3-rounds)
    &\addcheckemoji{}    & \addcrossemoji{}  \vspace{1pt}\\
    
    GPT-RE~\cite{DBLP:journals/corr/abs-2305-02105}    &\addcrossemoji{}    & \addcheckemoji{} &  \addcrossemoji{}
    &\addcheckemoji{}    & \addcrossemoji{}  \vspace{1pt}\\

    CoT-ER~\cite{DBLP:journals/corr/abs-2311-05922}    &\addcheckemoji{}  (3-agents)    & \addcrossemoji{} &  \addcheckemoji{} (3-rounds)
    &\addcheckemoji{}    & \addcrossemoji{}  \vspace{1pt}\\

    \midrule

    LM vs LM~\cite{DBLP:journals/corr/abs-2305-13281}    &\addcheckemoji{} (2-agents)   & \addcheckemoji{} &  \addcheckemoji{} (3-stages)
    &\addcheckemoji{}    & \addcrossemoji{}  \vspace{1pt}\\

    Multiagent Debate~\cite{DBLP:journals/corr/abs-2305-14325}    &\addcheckemoji{} (2-agents)   & \addcrossemoji{} &  \addcheckemoji{} (3-stages)
    &\addcheckemoji{}    & \addcrossemoji{}  \vspace{1pt}\\

    MAD~\cite{DBLP:journals/corr/abs-2305-19118}    &\addcheckemoji{} (3-agents)   & \addcheckemoji{} &  \addcheckemoji{}  (3-stages)
    &\addcheckemoji{}    & \addcrossemoji{}  \vspace{1pt}\\

    PRD~\cite{DBLP:journals/corr/abs-2307-02762}    &\addcheckemoji{} (2-agents)     & \addcheckemoji{} &  \addcheckemoji{}  (3-stages) 
    &\addcheckemoji{}    & \addcrossemoji{}  \vspace{1pt}\\

    SPP~\cite{DBLP:journals/corr/abs-2307-05300}    &\addcheckemoji{} (>3-agents)    & \addcheckemoji{} &  \addcheckemoji{} (4-stages) 
    &\addcheckemoji{}    & \addcrossemoji{}  \vspace{1pt}\\

    \textbf{Our CooperKGC}    &\addcheckemoji{}  (3-agents)
    & \addcheckemoji{} &  \addcheckemoji{} (3-stages)
    &\addcheckemoji{}    & \addcheckemoji{} (3-tasks) \vspace{1pt}\\
        
    \bottomrule
\end{tabular}
}

\caption{Comparison with previous methods. The upper half represents LLM-based KGC method, while the lower shows the emerging  multi-agent approach. }

\label{tab:comparison}
\end{table*}

\subsection{LLM-based Knowledge Graph Construction}

Recent years have witnessed a surge of interest in leveraging the remarkable advancements achieved by LLMs within the realm of KG. 
Notably, \cite{DBLP:journals/corr/abs-2305-13168} delves into the application of LLMs in KG construction and reasoning tasks.
Building on this foundation, \cite{DBLP:journals/corr/abs-2310-06671} integrates KG structural information into LLMs, employing self-supervised structural embedding pre-training.
\cite{DBLP:journals/corr/abs-2302-10205}, in a novel perspective, proposes a multi-turn question and answer architecture. 
Furthermore, \cite{DBLP:journals/corr/abs-2310-05066} pinpoints the unspecified task description as a key factor hindering the performance of contextual information extraction.
To address this, a guided learning framework is introduced to enhance the extraction model's alignment with specified guidelines.
Departing from the conventional isolation of KGC as a singular task, our approach advocates a departure from such isolation by fostering collaboration among a group of expert model agents in a multi-round social environment.

\subsection{Interactive Collaboration of Multiple Agents}
Recent developments highlight the effectiveness of collaborative efforts among language model agents, offering potential to enhance individual LLM capabilities. Various interaction architectures have emerged, assigning agents to specific roles. For example, setups like that in \cite{DBLP:journals/corr/abs-2305-14325} engage two agents in debates, enhancing factuality and reasoning albeit with increased computational costs. Similarly, \cite{DBLP:journals/corr/abs-2305-13281} introduces an examiner LLM to validate claims, uncovering factual errors through division of labor. Contributions such as the ChatLLM network \cite{DBLP:journals/corr/abs-2304-12998} foster dialogue and collective problem-solving among language models. Others introduce judges to summarize debates and provide conclusions \cite{DBLP:journals/corr/abs-2305-19118,DBLP:journals/corr/abs-2305-11595}, or frameworks incorporating peer review and discussion to address self-enhancement bias \cite{DBLP:journals/corr/abs-2307-02762}. Our study presents an interactive architecture tailored for a knowledge graph construction team, focusing on multiple information extractions and feedback compatibility.

\begin{figure*}[t]
    \includegraphics[width=1.0\textwidth]{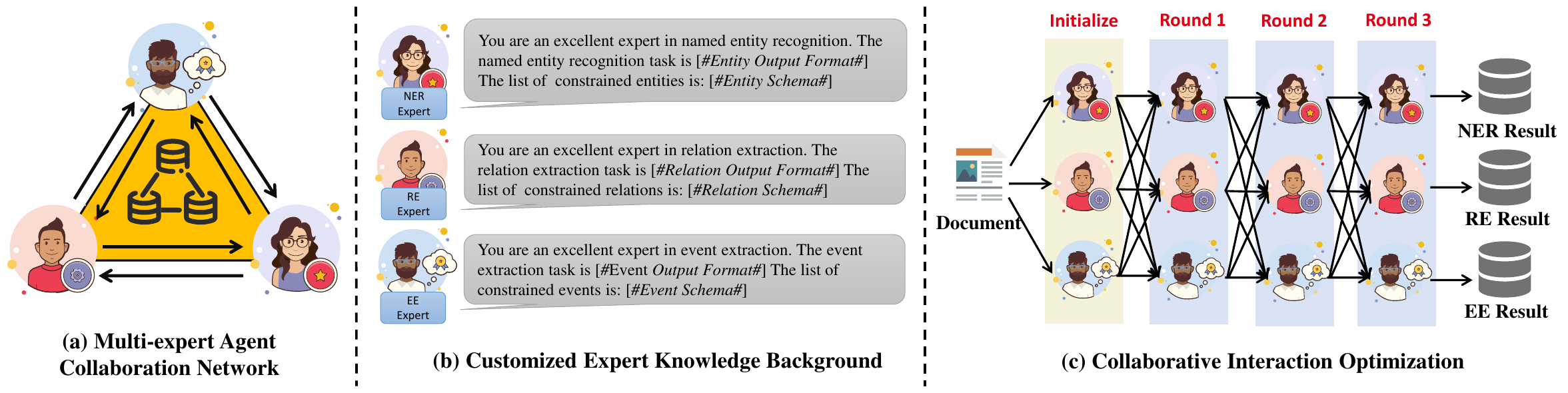}
    \caption{
     The overview of our CooperKGC.}
    \label{fig:method}
\end{figure*}

\section{Methodology}
\label{method_main}

Illustrated in Figure 1, we introduces a collaborative framework \textsc{CooperKGC}, aimed at advancing knowledge graph construction by concurrently extracting component elements such as entities, relations, and events.
Notably, our method could extend beyond the confines of the selected three tasks, offering flexibility through a dynamically formulated team collaboration network tailored to specific task requirements.  

\subsection{Construction of Multi-expert Agent Collaboration Network}
\label{sec:Agent-Collaboration-Network}

Traditional methods treat expert agents, each equipped with distinct back-ends, as isolated nodes within the collaborative network. 
These nodes independently contribute to task-solving through separate thinking chains, and a central adjudication node amalgamates and rectifies their responses. 
However, this conventional solution reveals two flaws:
(1) The adjudication node, functioning as the central hub, exhibits low fault tolerance and demands substantial reasoning ability to assimilate opinions from nodes spread across diverse collaborative networks;
(2) The team heavily relies on the ruling node as the sole consensus mechanism, hindering effective interactions between participants in the KGC task. 
In response to these limitations, we advocate a decentralized collaborative network communication scheme. 
Here, each expert agent backend, responsible for handling a specific task, establishes a bidirectional communication channel with any other expert agent backend. 
Despite the asynchronous nature of message production during practical operations, we adopt rounds as the fundamental unit of interaction to accomplish designated tasks and facilitate replica communication among expert agents.
It is noteworthy that, although our approach draws inspiration from the Byzantine Fault Tolerance \cite{DBLP:journals/toplas/LamportSP82} to form a distributed network, the message records held by each agent node differ.
In the process of replica communication, we implement message simplification, whereby extraction results complying with schema constraints are distilled.  
The formalization of the abstract collaboration network comprises three fundamental components:

\emph{\textbf{Expert Nodes}.} Expert nodes embody agents proficient in specific sub-tasks within KGC. They assimilate context from their peers at the preceding time step and formulate responses based on the input text $\mathcal{X}$. 
Notably, an Expert node can take various forms, including a vanilla LLM guided by explicit instructions, a self-reflective agent with a chain of thinking, or an agent explicitly leveraging domain knowledge through external knowledge bases or tool libraries.
With this foundation, our focus shifts to the collaborative functions between agents. 
Formally, the response $r^t_i$ of the $i$-th agent at the $t$-th round is expressed as a function $\mathcal{F}^i_t$, mapping from the base input text $\mathcal{X}$, prompt $p^i_t$, and predecessor expert agent’s replicas $\mathcal{R}_{t-1}$: 
$r^i_t = \mathcal{F}^i_t \left( \mathcal{X},p^t_i, \mathcal{R}_{t-1} \right) $, 
where $\mathcal{R}_{t-1} = \left\{ r_{t-1, j} | j = 1, 2, ...  \right\}$. 
 Let $\mathcal{A}$ be the set of all expert nodes and  $T$ be the maximum round.

\emph{\textbf{Communication Edges}.}
A two-way communication channel facilitates the exchange of insights among expert nodes $\mathcal{A}$ in the KGC collaboration network. 
In this context, we define $\mathcal{E}$ as the set encompassing all edges within the system. 
Recognizing the nuanced distinctions in information dissemination, we establish directional edges, represented by $e_{m,n} = (a^m_{t-1}, a^n_{t}) \in \mathcal{E}$,  where $a^m_{t-1}$ and $a^n_{t}$ signify the adjacent agent responsible for transmitting replica. It was, $a^n_{t}$ can perceive the replica passed from $a^m_{t-1}$ as its contextual input.
Thus, the expert nodes are intricately linked through these communication edges, constituting the interactive communication units  $\mathcal{C} = (\mathcal{A}, \mathcal{E})$ .

\emph{\textbf{Replicas Delivery}.} 
In the interactive communication unit $\mathcal{C}$,  replica delivery serves as the conduit guiding the flow of information from an agent in $(t-1)$-th round  to the input message queue of another agent in $t$-th round.
To streamline this intricate exchange, we designate a specific simplification function $\mathcal{S}$ to simple the the information: $d_{t-1}=\mathcal{S}(r_{t-1})$, where $\mathcal{S}$ predigest the complex CoT reasoning process.
Therefore, the replicas queue collected by the $i$-th expert node is expressed as
$\mathcal{D}^i = \left\{ d^j_{t-1} | j \neq i  \right\}$.

\subsection{\textbf{Customized Expert Knowledge Background}.}
\label{sec:Customized-Expert}

In order to unleash the ability of different expert agents to collaborate on complex extraction problems, we introduce customized expert knowledge background.
This context comprises three key components: 
(1) Opening statement $\mathcal{P}_o$, where each expert agents is presented with a directive elucidating how it can contribute its unique expertise to address a KGC task; 
(2) Task definition $\mathcal{P}_t$, which outlines the specifics of the knowledge graph extraction,
including the targeted elements and the guiding schema;
and (3) In-context demonstration $\mathcal{P}_c$, involving the selection of a limited set of $
\mathcal{M}$ instances. The overarching objective of this in-context demonstration is to furnish LLMs with illustrative examples.

\emph{\textbf{Opening Statement}.} 
As first part of the prompt, $\mathcal{P}_o$ contains a high-level instruction: 
\emph{"You are a knowledge graph constructor, need to synthesise relation extraction agent, named entity recognition agent, and event extraction agent to constitute an extraction collaborative team, which guides the agents to refine their results by referring to the extraction answers of others."}

\emph{\textbf{Task Definition}.} 
The task description $\mathcal{P}_t$ can be further decomposed into three components, as exemplified by the RE agent:(1) The first sentence of the task description, \emph{"You are an excellent expert in relation extraction."} is a constant that tells the LLM that it needs to focus on the relation extraction task;
(2) The second sentence defines the output format of the task: \emph{"Each result is returned as a tuple, e.g. [(head entity 1, relation type 1, tail entity 1), ...]"}. 
(3) The third sentence points to a specific list of relation types : \emph{"The list of constrained relations is: [\#Relation 1: [\#Head Entity Type 1, \#Tail Entity Type 2]...]"}.

\begin{algorithm}[t] 
  \caption{The Optimization Process of \textbf{CooperKGC}
  }  
  \label{algo:cooper}
  \KwIn{Input Text $\mathcal{X}$ , Expert Nodes $\mathcal{A}$, Communication Edges $\mathcal{E}$, Communication Unit $\mathcal{C}$, Round $\mathcal{N}$ } 
  \KwOut{KGC result $\mathcal{Y}^i$ for each $a_i \in \mathcal{A}$}

  \For{$a_i \in \mathcal{A}  $}{
        \Comment{Initial extraction results}
        $r^i_0 = \mathcal{F}^i_0\left( \mathcal{X} \| \mathcal{P}_o, \mathcal{P}_t,\mathcal{P}_c \right)  $;
        $d^i_0 = \mathcal{S}(r^i_0)$\;
  }
  \For{$ t = 1;\mathcal{N} $}{
    \For{$a_i \in \mathcal{A}  $}{
        \Comment{Replicas delivery by edges}
        $ \mathcal{D}^i_{t-1} \leftarrow $  $Transfer(e_{m,i})$,  $ \forall i, e_{m,i} = (a^m_{t-1}, a^i_{t}) \in \mathcal{E}$\;
        \Comment{Refine results by referring others}
        $r^t_i = \mathcal{F}^i_t\left( \mathcal{X} \| \mathcal{D}^i_{t-1}\| \mathcal{P}_v \|\mathcal{P}_o, \mathcal{P}_t,\mathcal{P}_c \right)  $;
        $d^i_t = \mathcal{S}(r^i_t)$\;
    }
}
     \Comment{Extract final answer, filter $d^i_t$ whose format does not comply with the constraints}
    $ \mathcal{Y}^i \leftarrow $ filter\_ans$(d^i_t | a^i_t \in \mathcal{A},\mathcal{C},\mathcal{X})$\;
\end{algorithm}

\emph{\textbf{In-context Demonstration}.} 
Some studies \cite{DBLP:journals/corr/abs-2305-02105,DBLP:journals/corr/abs-2311-05922,DBLP:conf/acl-deelio/LiuSZDCC22} show improvements in contextual learning by selecting few-shot demonstrations based on similarity. Our contextual prompts $\mathcal{P}_c$ are introduced as N-way K-shot sampling of the demonstration samples $\mathcal{M} = N \times K$, providing direct evidence about the task and references to predictions. 
However, limited by the input tokens of the LLMs, a single prompt may not contain all supported instances, so we use a sentence embedding similarity-based approach to select the $\mathcal{M}$ examples with the closest Euclidean distance as contexts.

\subsection{Collaborative Interaction Optimization}
\label{sec:optimization}

In the context of team collaboration optimization, the need for meticulous decomposition design diminishes, thus we reach to the periphery of the age-old adage, "Two heads are better than one."
As shown in Algorithm~\ref{algo:cooper}, after collecting replicas by other expert agents, we further provide collaboration prompts $\mathcal{P}_v$:
\emph{ The relation extraction answer you gave in the last round of collaboration was  
 "\#\#LAST\_ROUND\_RESULT\#\#". The answer given by the NER expert agent was "\#\#NER\_RESULT\#\#", The EE expert agent was "\#\#EE\_RESULT\#\#". You should refer to other members to revise your answer."}

\section{Experiments}

We conduct comprehensive experiments to evaluate the performance by answering the following research questions:
\begin{itemize}
    \item \textbf{RQ1}: How does our CooperKGC perform through teamwork when competing against SOTA?
    \item \textbf{RQ2}: What is the impact of the expert agents and the communication rounds in multi-round interactions in teamwork?  
    \item \textbf{RQ3}: How effective is the proposed CooperKGC in extracting different types of entities, relations and events?
\end{itemize}

\subsection{Experiment Settings}

\subsubsection{\textbf{Dataset.}} 

As to the NER task, we conduct experiments  on the following popular benchmark: \textbf{Conllpp} \cite{DBLP:conf/emnlp/WangSLLLH19},
\textbf{OntoNotes5.0} \cite{DBLP:conf/conll/PradhanMXNBUZZ13} and  \textbf{MSRA} \cite{DBLP:conf/acl-sighan/Levow06}. 
For RE task, we conduct experiments  on the following popular benchmark: \textbf{NYT11-HRL} \cite{DBLP:conf/aaai/TakanobuZLH19}, 
\textbf{Re-TACRED} \cite{DBLP:conf/aaai/StoicaPP21}, 
and  \textbf{DuIE2.0} \cite{DBLP:conf/nlpcc/LiHSJLJZLZ19}.
For EE task, there are two standard datasets:   \textbf{ACE05} \cite{ace2005-annotation} and  \textbf{DuEE1.0} \cite{DBLP:conf/nlpcc/LiLPCPWLZ20}.

\subsubsection{\textbf{Baselines.}} 
In our experimental framework, we opt for \textbf{AutoKG} \cite{DBLP:journals/corr/abs-2305-13168} as the implementation of Vanilla LLMs for KGC realm, which defines an end-to-end extraction workflow  through the manual templates. 
Expanding on this foundation, \textbf{ChatIE} \cite{DBLP:journals/corr/abs-2302-10205} refines the extraction process using a two-round method. 
Taking RE as an example, this method entails the initial extraction of the relation, followed by the output of the associated entity span. 
This sequential approach mirrors a cognitive model's thought process, explicitly delineating the steps of task decomposition. 
Further, \textbf{CoT-ER} \cite{DBLP:journals/corr/abs-2311-05922} introduces an explicit evidence reasoning method, characterized by three rounds of processing. 
In the first and second rounds, the LLM is required to output concept-level entities corresponding to head and tail entities.
Subsequently, in the third round, the extraction of relevant entity spans occurs, establishing a specific relationship between these two entities with explicit evidence.


\subsection{Performance Comparison with SOTA (RQ1)}

\begin{table}[t]
\caption{
F1-score results for 3 KGC tasks (NER, RE, EE) on the 8 datasets.}
\centering
\setlength{\tabcolsep}{4pt}
\resizebox{1.0\textwidth}{!}{
\begin{tabular}{l|ccc|ccc|cc}
    \toprule
    \multirow{2}{*}{\textbf{Model}} & \multicolumn{3}{c|}{\textbf{NER}} & \multicolumn{3}{c|}{\textbf{RE}} & \multicolumn{2}{c}{\textbf{EE}}  \\
    \cline{2-9}
    & Conllp & OntoNotes5.0 & MSRA & NYT11-HRL & RE-TACRED & DUIE2.0 & ACE05 & DUEE1.0 \\
    \midrule
AutoKG (0-shot) & 50.6 & 40.4 & 56.8 & 12.5 &17.2 & 26.9 & 20.7 & 68.7 \\
ChatIE (0-shot) & 58.4 & 47.5 & \underline{57.7} & 37.5 & 43.9 & 68.4 & 29.7 & 72.0 \\
CoT-ER (0-shot) & \underline{60.1} & \underline{52.6} & 57.3 & \underline{45.3} & \underline{44.2} & \underline{68.7} & \underline{43.1} & \underline{73.1} \\
    \midrule
AutoKG (1-shot) & 55.3 & 40.9 & 56.8 & 26.5 &22.5 & 43.6 & 26.9 & 71.2 \\
ChatIE (1-shot) & \underline{61.3} & 49.2 & \underline{59.2} & 44.7 & 47.5 & 70.2 & 31.2 & \underline{74.2} \\
CoT-ER (1-shot) & 61.1 & \underline{53.7} & 58.7 & \underline{47.4} & \underline{48.3} & \underline{71.5} & \underline{45.3} & 74.1 \\
    \midrule
CooperKGC (0-shot) & \textbf{61.3}\color{red}{(+10.7)} & \textbf{53.8}\color{red}{(+13.4)} & \textbf{60.2}\color{red}{(+3.4)} & \textbf{45.7}\color{red}{(+33.2)} & \textbf{47.1}\color{red}{(+29.9)} & \textbf{72.2}\color{red}{(+45.3)} & \textbf{47.2}\color{red}{(+26.5)} & \textbf{79.5}\color{red}{(+10.8)} \\
CooperKGC (1-shot) & \textbf{61.5}\color{red}{(+6.2)} & \textbf{55.4}\color{red}{(+14.5)} & \textbf{60.9}\color{red}{(+4.1)} & \textbf{49.2}\color{red}{(+22.7)} & \textbf{51.2}\color{red}{(+28.7)} & \textbf{73.6}\color{red}{(+30.0)} & \textbf{47.5}\color{red}{(+20.6)} & \textbf{81.3}\color{red}{(+10.1)} \\
\bottomrule
\end{tabular}
}
\label{table:performance}
\end{table}

Our study conducts comprehensive experiments on 0-shot and 1-shot settings across 8 datasets, each with 100 samples from test/valid sets, evaluating results using micro F1. We use the "gpt-3.5-turbo" API for both baseline models and proposed methods, with a temperature parameter set to 0.0 and average results reported over three runs. Our method sets a maximum of 4 rounds and 3 KGC team members.
For the English dataset, the default customized expert knowledge background for the NER task is based on Conllpp, the RE task is NYT11-HRL, and the EE task is ACE05. 
Similarly, for the Chinese dataset, the NER task is based on MSRA, the RE task is DUIE2.0, and the EE task is DUIE1.0. 
Table~\ref{table:performance} presents F1-score results for 3 KGC tasks across datasets, revealing:

\textbf{(1) CooperKGC enhances overall performance across diverse tasks.} Compared to the vanilla method, CooperKGC shows significant improvements in both 0-shot and 1-shot settings. 
In contrast, for a simple extraction approach like AutoKG with a single round of LLM calls, on the one hand, the overly heavy information input for task comprehension and rule constraints poses a challenge for a single model. 
On the other hand there is a lack of sufficient inference steps for a self-debugging process. Our approach, multiple rounds of interactions alleviate this anxiety of requiring "hit-and-miss" reasoning, making it easier to explicitly identify erroneous intermediate feedbacks during the interactions.

\textbf{(2) Teamwork is an effective implicit reasoning chain.}
Taking the NYT11-HRL dataset as an example, although ChatIE improved by 25.0 over the baseline in the 0-shot setting while achieving an improvement of 18.2 over the baseline in the 1-shot setting, we believe that the gain stems from decomposing the extraction process into two phases. 
Among the first stage is determining the types of relations involved in a given sentence, which often involves multiple relations in a single sentence. 
The second stage then designs triple extraction templates for each relation, which clearly indicates the sub-tasks to be accomplished in each stage.
CoT-ER uses head-to-tail mapping to induce LLM to generate explicit evidence of reasoning, resulting in an improvement of 32.8 over the baseline in the 0-shot setting.
CooperKGC outperforms both, with a 33.2 improvement in the 0-shot and a 22.7 improvement in the 1-shot setting. 
We believe that building collaborative teams contributes to "\emph{Brain Storming}" \cite{osborn1953applied}, where each round of the brainstorming process is performed by the members of team. 
By collecting evidence from other members in each round of interactions,  agent's responses is fine-tuned from the previous round. 
Although there is no reasoning path, this proactive optimisation shows more encouraging prospects than passive methods.

\subsection{Analysis of Team Members and Interaction Rounds (RQ2)}

\begin{figure*}[t]
    \includegraphics[width=1.0\textwidth]{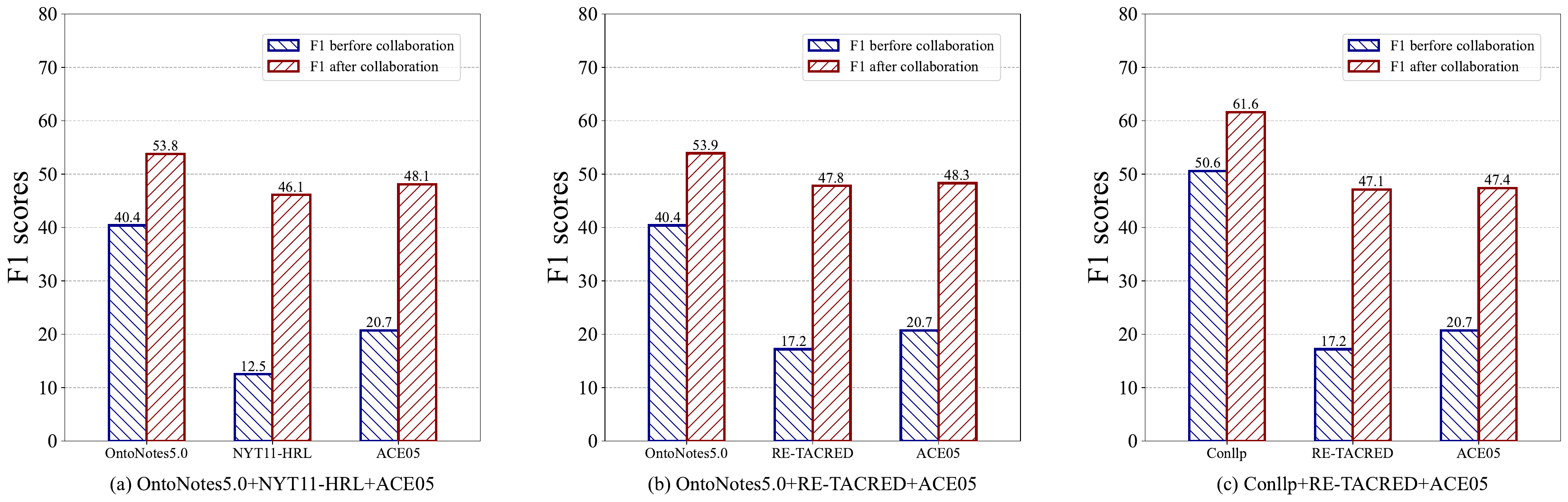}
    \caption{
    Equipping KGC agents with different expert knowledge backgrounds.}
    \label{fig:combination}
\end{figure*}

To further investigate the impact brought by the combination of intelligences with different expert knowledge backgrounds on team collaboration, we introduce an experiment to analyse the diverse combination of team members.
Specifically, We experiment on 0-shot setting and the number of team members is fixed to 3. By replacing the expert knowledge backgrounds representing NER agent, RE agent, and EE agent, we analyse which kind of expert knowledge backgrounds (mainly the schema constraints in the task description $\mathcal{P}_t$) could produce better benefits for the team construction goals. 
Figure~\ref{fig:combination} shows the results of equipping KGC agents with different expert knowledge backgrounds, and we observe that \texttt{combination b (OntoNotes5.0+RE-TACRED+ACE05)} allows EE expert agents to achieve the best extraction performance, and the richer variety of relation types guided by RE-TACRED allows EE agents to discover more potential arguments compared with \texttt{combination a}. 
In addition, \texttt{combination b} achieves a more comprehensive improvement compared to \texttt{combination} \texttt{c}. 
We analyse the schema of OntoNotes5.0 versus Conllp and find that three of the entity categories are the same ("\emph{PER}", "\emph{LOC}", "\emph{ORG}"), while the remaining 15 more specialised entity categories refine the "\emph{MISC}" category in Conllp, which results in benefits in extraction performance for the RE-TACRED and ACE05 datasets.
We therefore conclude that more specialised expert agents, i.e., equipped with fine-grained schema constraints, can bring more insightful information to guide teamwork.

\begin{table}[t]
\setlength{\tabcolsep}{8pt}
\caption{
Micro-F1 Performance under different member assignments.}
\centering
\resizebox{0.65\textwidth}{!}{
\begin{tabular}{l|ccc}
    \toprule

Team Members & Conllp  & NYT11-HRL  & ACE05 \\
\midrule
3-Agent & \colorbox{Mycolor2}{61.3}  & \colorbox{Mycolor2}{45.7}  & \colorbox{Mycolor3}{47.2} \\
3-Agent + \textsc{ontoNotes} & \colorbox{Mycolor3}{58.4}  & \colorbox{Mycolor1}{\textbf{46.3}}  &  \colorbox{Mycolor2}{48.3} \\
3-Agent + \textsc{re-tacred} & \colorbox{Mycolor1}{\textbf{62.2}}  & \colorbox{Mycolor3}{38.4}  & 47.4 \\
3-Agent + \textsc{both} & 58.6  & 38.9  & \colorbox{Mycolor1}{\textbf{48.4}} \\
\midrule
3-Agent \textsc{(all conllp)} & 60.8  & -  & - \\
3-Agent \textsc{(all nyt-hrl)} & -  & 44.9  & - \\
3-Agent \textsc{(all ace05)} & -  & -  & 29.1 \\
\bottomrule

\end{tabular}
}
\label{Tab:Team-member}
\end{table}

\begin{figure}[t]
    \centering
    \includegraphics[width=0.65\textwidth]{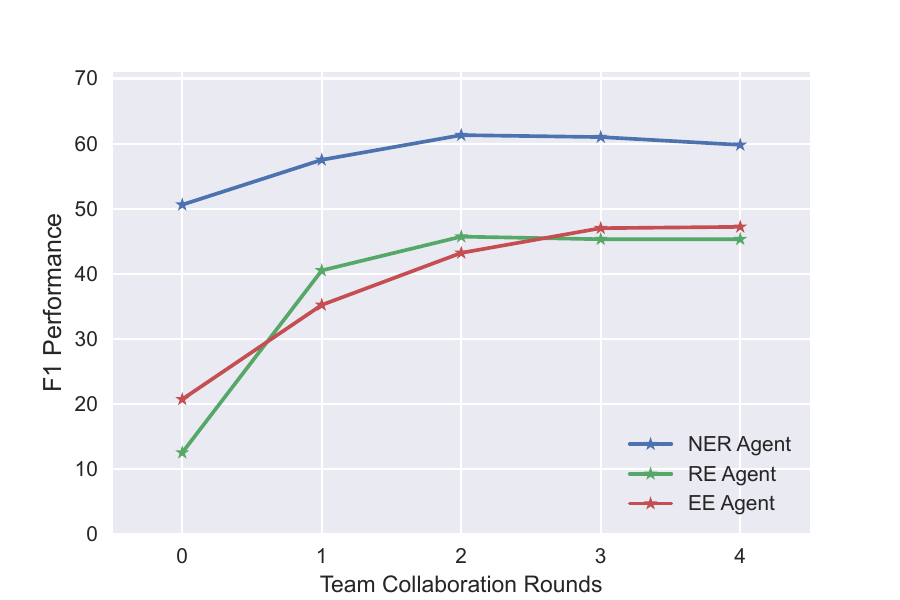}
    \caption{
   Micro-F1 Performance with Increased Rounds.}
    \label{fig:round}
\end{figure}

\begin{table*}[th]

\caption{A qualitative example of CooperKGC on expert agent collaboration.}
\resizebox{1.0\textwidth}{!}{
 \begin{tabular}{p{0.3cm}|p{3cm}|p{7cm}|p{11cm}}

  \toprule
  \textbf{R} & \textbf{NER Agent} & \textbf{RE Agent} & \textbf{EE Agent}\\
  \midrule
    0 & (LOC, Palestinian section of the border crossing) 
    & (Palestinian section of the border crossing, location-located\_in, Israeli troops)
   (Israeli troops, person-nationality, Palestinians)
    & [\{Trigger Type: Conflict-Attack, \textcolor{reds}{\textbf{Trigger Word: taken over}}, Arguments: (Attacker, Israeli troops),\textcolor{reds}{\textbf{(Target: Palestinian section of the border crossing)}}\}, \{Trigger Type: Movement:Transport, Trigger Word: return, Arguments: (Destination, the Palestinian section of the border crossing)\}]
    \\
   \midrule
    1 & \textcolor{blues}{\textbf{(PER, Israeli troops)}}, \textcolor{blues}{\textbf{(LOC, border)}}, \textcolor{blues}{\textbf{(ORG, police)}}
    & \textcolor{blues}{\textbf{(Israeli troops, location-located\_in, the Palestinian section of the border crossing)}}, 
    \textcolor{blues}{\textbf{(Israeli troops, person-nationality, Israeli)}},
    \textcolor{blues}{\textbf{(Six Palestinian police officers, person-nationality, Palestinians)}} 
    & [\{Trigger Type: Conflict-Attack, Trigger Word: uprising, Arguments: (Attacker, Israeli troops), \textcolor{reds}{\textbf{(Place, the Palestinian section of the border crossing)}}\}, \{Trigger Type: Movement:Transport, Trigger Word: return, Arguments: (Destination, the Palestinian section of the border crossing)\}]
      \\
   \midrule
    2 &  (PER, Israeli troops), (LOC, border), (ORG, police), \textcolor{blues}{\textbf{(PER, Six Palestinian police officers)}}
    &  \textcolor{blues}{\textbf{(Israeli troops, person-place\_lived, the Palestinian section of the border crossing)}}, (Israeli troops, person-nationality, Israeli),
    (Six Palestinian police officers, person-nationality, Palestinians)    
    &  [\{Trigger Type: Conflict-Attack, Trigger Word: uprising, Arguments: (Attacker, Israeli troops), (Place, Israeli)\}, \{Trigger Type: Movement:Transport, Trigger Word: return, Arguments: \textcolor{reds}{\textbf{(Destination, border)}}, \textcolor{reds}{\textbf{(Artifact, Israeli troops)}}\}]
      \\
   \midrule
    3 & (PER, Israeli troops), (LOC, border), (ORG, police), (PER, Six Palestinian police officers)
    & (Israeli troops, person-place\_lived, the Palestinian section of the border crossing), (Six Palestinian police officers, person-nationality, Palestinians)
    &  [\{Trigger Type: Conflict-Attack, Trigger Word: uprising, Arguments: (Attacker, Israeli troops), (Place, Israeli)\}, \{Trigger Type: Movement:Transport, Trigger Word: return, Arguments: \textcolor{reds}{\textbf{(Destination, border)}}, (Artifact, Six Palestinian police officers)\}]
      \\
   \midrule
    4 & (PER, Israeli troops), (LOC, border), (ORG, police), (PER, Six Palestinian police officers)
    & (Israeli troops, person-place\_lived, the Palestinian section of the border crossing),  (Six Palestinian police officers, person-nationality, Palestinians)
    & [\{Trigger Type: Conflict-Attack, Trigger Word: uprising, Arguments: (Attacker, Israeli troops), (Place, Israeli)\}, \{Trigger Type: Movement:Transport, Trigger Word: return, Arguments: (Destination, the Palestinian section of the border crossing), (Artifact, Six Palestinian police officers)\}] \\

\bottomrule
 \end{tabular}
}
\label{Tab:case}
\end{table*}

Another question is whether it is possible to equip with more agents to make more gains for our team. 
Table~\ref{Tab:Team-member} shows the results of both kinds of experiments, the upper one is to add additional agents to the original team, and the results show that \texttt{Team (3-Agent+BOTH)} makes the extraction results of ACE05 improved by adding a NER Agent and a RE Agent. 
However, another risk is also demonstrated, in both \texttt{Team (3-Agent+OntoNotes)} and \texttt{Team (3-Agent+RE-TACRED)} it is observed that when more authoritative expert agents are introduced, it leads to a decrease in the extraction results of the agent for the same task, and this kind of unconscious opinion conformity is consistent with the concept of "\emph{Presentation of Self}" \cite{goffman2002presentation} in sociology.
In addition, inspired by "\emph{Self-consistency}" \cite{DBLP:conf/iclr/0002WSLCNCZ23}, in the bottom of Table~\ref{Tab:Team-member}  we explore the difference between the performance of the self-consistent voting method and CooperKGC on a single task.  Although the consistency method to some degree mitigates the randomness of the single agent producing the hallucinatory fact, it is nevertheless weaker than our results on all 3 representative datasets.
We argue that a single perspective is unable to access the interactive information provided by other experts, and thus suffers from "\emph{Information Cocoons}"\cite{sunstein2006infotopia}.

Next, we provide an analysis of the impact of the number of collaboration rounds on multi-agent teams. 
In Figure~\ref{fig:round}, we increase the number of rounds for interaction between agents while fixing the number of agents to 3. 
We find that the performance of the algorithm also increases with the number of collaboration rounds in the first 2 rounds on all three types of tasks.
However, the NER agent performance achieves its best in round 2, the RE agent in round 3, and additional collaboration by the EE agent over 3 rounds leads to a final performance similar to 3 rounds
collaboration.
Therefore, we believe that for tasks with simple extraction structures, too many interactions may lead to the introduction of undesirable hallucinations, hence a balance between performance and collaboration costs needs to be achieved on a task-specific basis.

\subsection{Case Study of Collaboration Process (RQ3)}

To illustrate the effectiveness of our proposed CooperKGC for collaborative interactions in KGC teams, Table~\ref{Tab:case} provides a qualitative example demonstrating the intermediate process.
Note the CoT reflection process such as "\emph{After considering the extraction results of other agents...}" is skipped, and the input sentence is an example of EE task "\textit{Six Palestinian police officers were allowed to return to the Palestinian section of the border crossing, which had been taken over by Israeli troops shortly after the start of the uprising.} "
we compare the results of the EE agent with the groundtruth, while the results of the NER agent and the RE agent are only for reference since there is no groundtruth.
The observations are as follows:
(1) \textbf{Knowledge Selection.} In Round 2, the EE agent borrows the \emph{LOC} entity "\emph{border}" newly discovered by the NER agent in the previous round and adds an argument \emph{(Destination, border)} to the original answer; 
(2) \textbf{Knowledge Correction.} In the 1st round of interactions, the EE agent corrects the wrong trigger word "taken over", which indicates that the team members have the ability to provide self-feedback;
(3) \textbf{Knowledge Aggregation.} Although the EE agent puts a wrong argument \emph{(Destination, border)} in round 3, it rectifies the hallucination facts generated in the interim by eliciting LLM semantic comprehension during the interaction.

\section{Conclusion and Future Work}

In this study, we initiated the formation of a KGC team by aggregating agents with diverse expertise.  
Our results highlight the collaborative potential of LLM agents, showcasing how agent networks can enhance task performance collectively.  
The emergence of human-like behaviors in collaboration aligns with sociological theories, leading to improvements in factuality, knowledge integration, and intellectual reasoning.  Future research could draw insights from sociologically derived architectures, expanding the application of CooperKGC variants to solve diverse collaborative tasks.

%
%
\bibliographystyle{splncs04}
\bibliography{mybibliography}

\end{document}